\documentclass[letterpaper, 10 pt, conference]{ieeeconf}

\IEEEoverridecommandlockouts                              

\usepackage{times}
\usepackage{epsfig}
\usepackage{graphicx}
\usepackage{amsmath}
\usepackage{amssymb}
\newcommand{\mypar}[1]{\vspace{0.1cm}\noindent\textbf{#1}}
\usepackage{multirow}
\usepackage{booktabs}
 \usepackage{hhline}
 \usepackage{colortbl}
 \usepackage{graphics} % for pdf, bitmapped graphics files
\usepackage{mathptmx} % assumes new font selection scheme installed
\usepackage{makecell}
\usepackage[T1]{fontenc}
\usepackage{bbding}
\usepackage{wasysym}

\usepackage{colortbl}
\usepackage{mathrsfs}
\usepackage{booktabs}
\usepackage{footmisc}
\usepackage{tablefootnote}
\usepackage{makecell}
\usepackage{algorithm}
\usepackage{algpseudocode}
\algnewcommand{\IIf}[1]{\State\algorithmicif\ #1\ \algorithmicthen}
\algnewcommand{\EndIIf}{\unskip\ \algorithmicend\ \algorithmicif}
\usepackage{algcompatible}

\let\labelindent\relax
\usepackage{enumitem}

\usepackage{amssymb}% http://ctan.org/pkg/amssymb
\usepackage{pifont}% http://ctan.org/pkg/pifont
\usepackage{balance}

\usepackage{footnote}

\makeatletter
\let\NAT@parse\undefined
\makeatother
\usepackage[colorlinks,linkcolor=red,anchorcolor=blue,urlcolor=magenta,citecolor=blue]{hyperref}

\usepackage[capitalize]{cleveref}
\crefname{section}{Sec.}{Secs.}
\Crefname{section}{Section}{Sections}
\Crefname{table}{Table}{Tables}
\crefname{table}{Tab.}{Tabs.}

\title{\LARGE \bf
TransDARC: Transformer-based Driver Activity Recognition with Latent Space Feature Calibration
}
\author{Kunyu Peng, Alina Roitberg, Kailun Yang, Jiaming Zhang, and Rainer Stiefelhagen
\\Institute for Anthropomatics and Robotics, Karlsruhe Institute of Technology
\\  {\tt\small \{firstname.lastname\}@kit.edu}
\thanks{Acknowledgement: The research leading to these results was supported by the SmartAge project sponsored by the Carl Zeiss Stiftung (P2019-01-003; 2021-2026) and the Competence Center Karlsruhe for AI Systems Engineering (CC-KING) sponsored by the Ministry of Economic Affairs, Labour and Housing Baden-Württemberg.
The authors would like to thank the consortium for the successful cooperation.}
}

\begin{document}

\maketitle
\thispagestyle{empty}
\pagestyle{empty}

\begin{abstract}
Traditional video-based human activity recognition has experienced remarkable progress linked to the rise of deep learning, but this effect was slower as it comes to the downstream task of driver behavior understanding. Understanding the situation inside the vehicle cabin is essential for Advanced Driving Assistant System (ADAS) as it enables identifying distraction, predicting driver’s intent and leads to more convenient human-vehicle interaction. At the same time, driver observation systems face substantial obstacles as they need to capture different granularities of driver states, while the complexity of such secondary activities grows with the rising automation and increased driver freedom. Furthermore, a model is rarely deployed under conditions identical to the ones in the training set, as sensor placements and types vary from vehicle to vehicle, constituting a substantial obstacle for real-life deployment of data-driven models. In this work, we present a novel vision-based framework for recognizing secondary driver behaviours based on visual transformers  and an additional augmented feature distribution calibration module. This module operates in the latent feature-space enriching and diversifying the training set at feature-level in order to improve  generalization to novel data appearances, (\textit{e.g.}, sensor changes) and general feature quality. Our framework consistently leads to better recognition rates, surpassing previous state-of-the-art  results of the public Drive\&Act benchmark on all granularity levels. Our code is publicly available at \url{https://github.com/KPeng9510/TransDARC}.
\end{abstract}

\section{Introduction}

Daily lives have clearly benefited from the rise of the automobile industry, \textit{e.g.}, through the reduced travelling time  and strengthened  connection between different countries, but everything has its double face.
According to the World Health Organization (WHO), around $2.2\%$ of total number of death -- $1.35$ million, were caused by traffic accidents in 2020~\cite{world2019world}.
The majority of such tragedies involve driver being engaged in distractive secondary activities, \textit{e.g.}, eating, drinking, having a call, or reading  and 36\% of such accidents could be avoided if no distraction occurred~\cite{dingus2016driver}.
Even in the case of highly automated driving, studies suggest that driver being engaged in certain behaviours such as interacting with the infotainment unit, negatively impact the  readiness-to-take-over the vehicle control~\cite{deo2019looking}.

Accurate Advanced Driving Assistant System (ADAS)~\cite{rommerskirchen2014impact} have strong potential to counter this issue by detecting such distractions and forecasting the risk of traffic accident at early time.
\begin{figure}[t]
\begin{center}
\includegraphics[width = .5\textwidth]{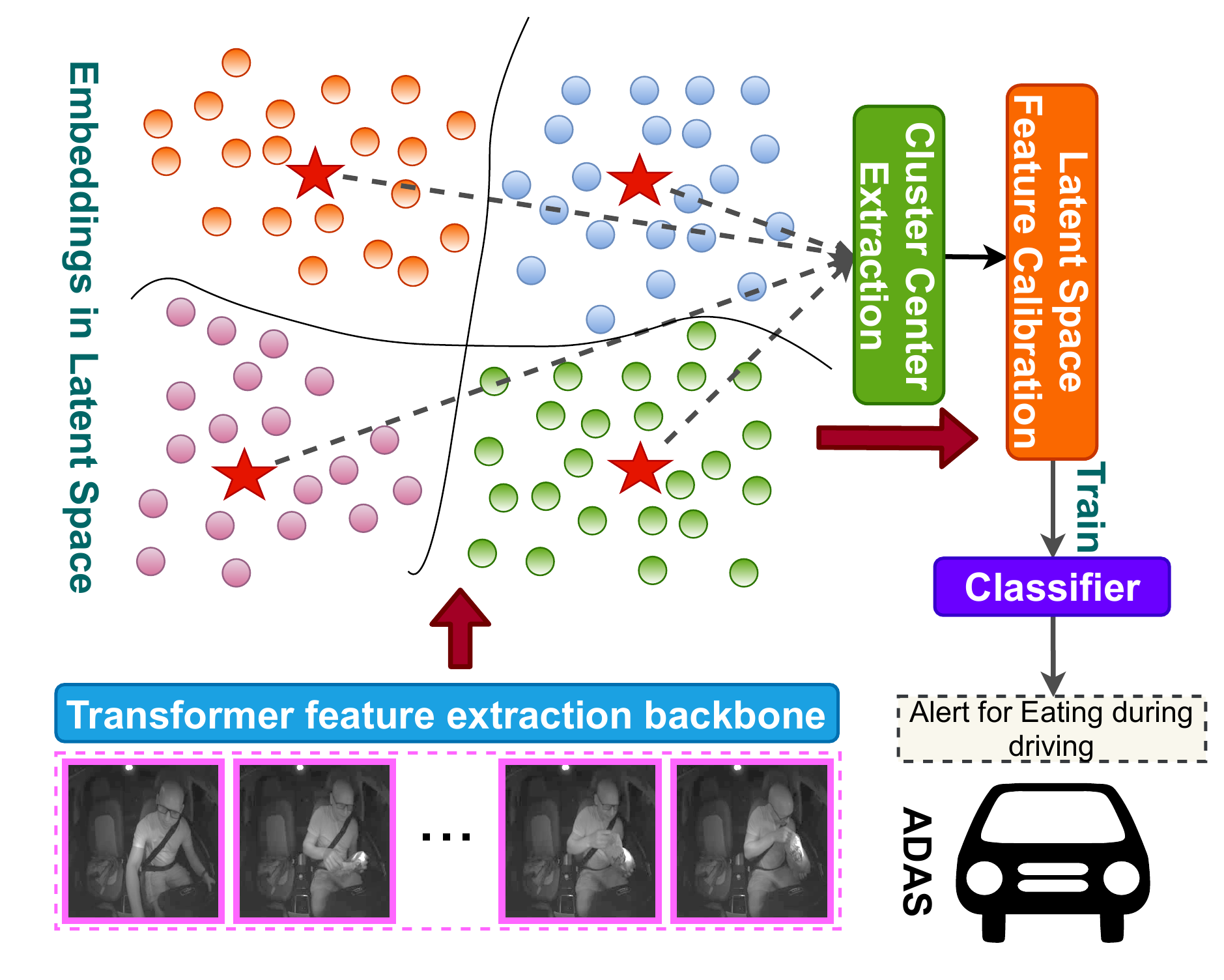}
\vskip-3ex
\caption{
An overview of the proposed \textsc{TransDARC} framework. We first train our transformer-based encoder for feature extraction and then use the statistic center selected differently according to its rareness to augment and generate new samples in latent space using the proposed augmented feature calibration.
In the second stage, an additional attention-based classifier is trained based on the generated calibrated training feature set in latent space.
The latent space calibration aims to improve the accuracy of driver activity recognition on not only for commonly existed classes, but also for rarely existed classes, while simultaneously considering the essential cross-task and -modality generalization ability, respectively.
}
\vskip-6ex
\label{fig:first}
\end{center}
\end{figure} 
\begin{figure*}[t]
\begin{center}
    \includegraphics[width = 1\textwidth]{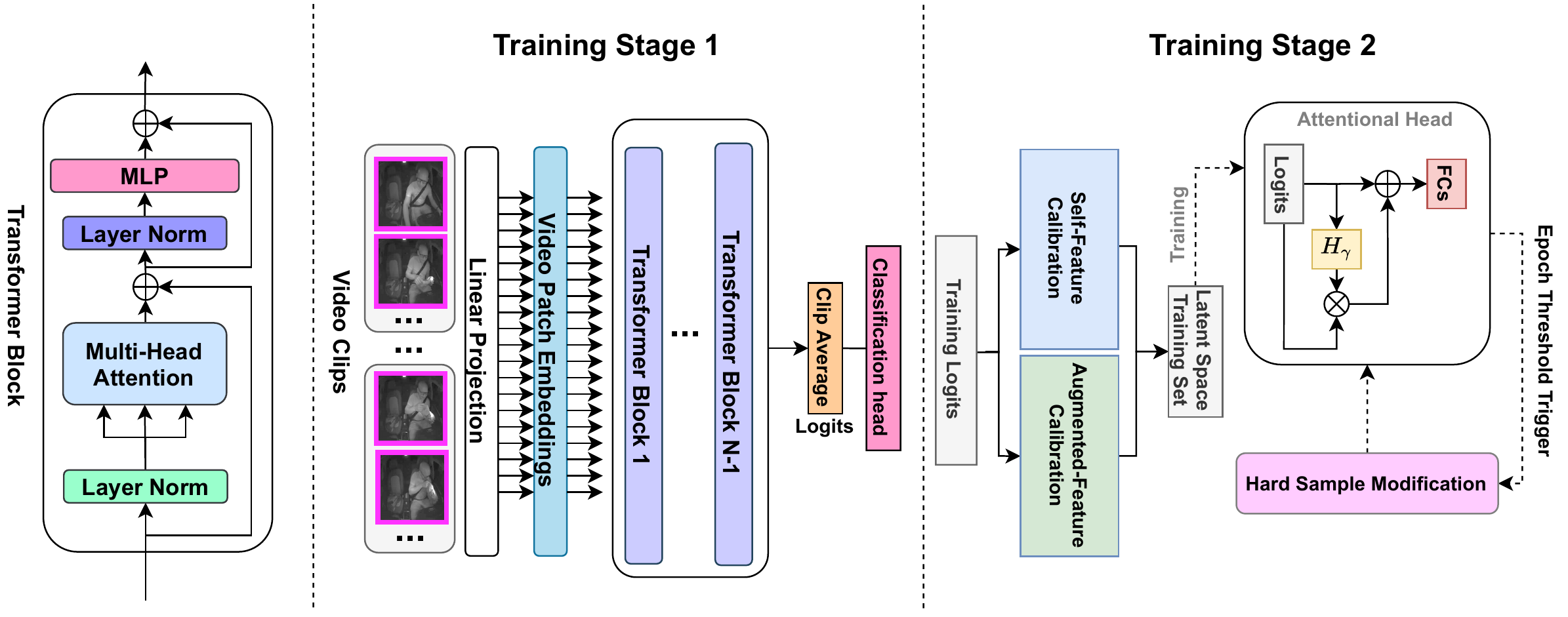}
\vskip-2ex
\caption{
A detailed introduction of the proposed \textsc{TransDARC} framework, which is composed of \textbf{Trans}former-based \textbf{D}river \textbf{A}ctivity \textbf{R}ecognition training stage and an additional latent space feature \textbf{C}alibration stage. Video Swin Transformer~\cite{liu2021video_swin} is selected as the transformer-based feature extraction backbone in our work. In the second stage, we use both self-feature calibration and augmented feature calibration approaches together with an additional attention-based classification head, where the attention $H_{\theta}$ is learned through stacks of fully connected layers.
}
\vskip-6ex

\label{fig:main}
\end{center}

\end{figure*} 
Automatic recognition of secondary driver activities can be viewed as a fine-grained downstream task of general video classification, where frameworks are often derived from existing approaches for standard activity recognition based,  \textit{e.g.}, on Convolutional Neural Networks (CNNs)~\cite{carreira2017quo,martin2019drive} and Graph Neural Networks (GNNs)~\cite{martin2019drive,yan2018spatial}.
However, existing driver activity recognition research indicates that there is still a long way to go for an accurate driver assistance~\cite{martin2019drive, wharton2021coarse, roitberg2020cnn_spatialtemporal}.
The recognition rates are especially low for (1) changes in data appearance (due to the sensor type or placement) and (2) for categories underrepresented in the training set.
Models which generalize well across different data domains are vital considering the diversity of inner vehicle structures  and different potential sensor placements.
The second aspect is also highly relevant in real-life applications, where the dataset categories are often unevenly distributed for practical reasons~\cite{martin2019drive} and recognition biases towards the most common categories constitute a significant issue.

In this work, we aim to utilize the recently emerged attention-based approaches for visual recognition in video~\cite{liu2021video_swin,arnab2021vivit,zhang2021token_shift,wang2021efficient} and present a novel vision-based framework for recognizing secondary driver behaviours based on visual transformers and  augmented feature distribution calibration.
For the initial feature extraction, we leverage the Video Swin Transformer~\cite{liu2021video_swin} to improve  the overall feature quality (which is usually done with CNNs or GNNs in driver observation~\cite{martin2019drive,roitberg2020cnn_spatialtemporal,martin2018body_pose,martin2020dynamic_interaction}).
To meet the previously mentioned challenges of generalization to novel data appearances and uneven distribution of driver behaviours during training, we equip the backbone  with the proposed feature calibration approach operating in the latent feature-space and  diversifying the training set at feature-level.
By assuming the distribution of each channel as Gaussian distribution in the latent space, more data points can be generated through the feature-level interpolation between existing vectors and the statistics of the different cluster centers.
The training set is thereby enriched in the high dimensional latent space, balancing the ratios among different driver behaviour types and increasing the generalization to new sensor setups.
We demonstrate the effectiveness of our approach on the public large-scale Drive\&Act dataset~\cite{martin2019drive}, compared to the previously published approaches and our implemented Video Swin-based~\cite{liu2021video_swin} baseline without additional feature calibration.
Our framework consistently outperforms  previously published approaches and baselines on all Drive\&Act tasks, surpassing the best previously published approach~\cite{wharton2021coarse} by a significant margin, (\textit{e.g.}, almost $25\%$ in the fine-grained subtask).
Our approach is especially effective in cases of data appearance changes, which is critical in real-life driving applications.

The contributions of this work are summarized as follows:
\begin{itemize}
    \item We for the first time address the challenging driver activity recognition task using visual  \textit{transformers}  instead of CNN- or GCN-based feature extraction approaches commonly used in driver observation~\cite{martin2019drive}. 
    The specific focus on long-term information aggregation in transformers~\cite{vaswani2017attention} makes this type of models especially suitable for learning driver behaviour representations, which is validated through our extensive experiments.
    \item We propose a novel  feature distribution calibration module operating in the latent space and using feature-level interpolations among different characterized feature clusters to enrich the training set as shown in Fig.~\ref{fig:first}. This module improves the feature quality and specifically encourages generalizability in the cross-modal setting   by diversifying the training set at feature-level. 
    We refer to our proposed \textsc{\textbf{Trans}}former-based \textsc{\textbf{D}}river \textsc{\textbf{A}}ctivity \textsc{\textbf{R}}ecognition with Latent Space Feature \textsc{\textbf{C}}alibration framework as  \textsc{TransDARC}.
    \item Our \textsc{TransDARC} model consistently outperforms all previously published  approaches by a large  margin on different driver observation tasks, including the fine-grained- and coarse driver activity recognition as well as human-object interaction  estimation, setting a new state-of-the-art on the public Drive\&Act benchmark. 
    The performance gain using \textsc{TransDARC} is especially high for underrepresented driver behaviours  and under cross-modality conditions.
\end{itemize}

\section{Related Works}
\noindent\textbf{Driver activity recognition.}
Traditional driver behavior recognition systems often rely on a manual feature construction process followed by a classification module like SVMs~\cite{ohn2014head_eye_hand} and random forests~\cite{xu2014realtime_random_forests}. The extracted feature vectors originate from hand- and body poses~\cite{martin2018body_pose,das2015performance_evaluation_hand}, eye-related inputs like driver gaze~\cite{zheng2015eye_gaze,braunagel2015driver_conditionally}, head patterns~\cite{ohn2014head_eye_hand,braunagel2015driver_conditionally}, as well as foot dynamics~\cite{rangesh2019forced_foot}.
Object recognition cues~\cite{weyers2019action_object} and physiological signals~\cite{bi2016queuing_eeg, cui2022eeg} are also associated for driver behavior observation.

With the prosperity of Convolutional Neural Networks (CNNs) in computer vision, the CNN-based deep learning pipeline gains popularity in a broad range of fields~\cite{he2016resnet}. 
Top-scoring CNNs~\cite{he2016resnet,wang2018nonlocal} and spatial temporal architectures like I3D~\cite{carreira2017quo} and P3D~\cite{qiu2017learning} are applied in driver activity analysis tasks~\cite{roitberg2020cnn_spatialtemporal}.
Trajectory feature~\cite{cheung2018identifying_trajectory_features} and optical flow~\cite{gebert2019endtoend_opticalflow} are also frequently explored to enhance driver behavior identification towards safe transportation.
Behera~\textit{et al.}~\cite{behera2020deep_cnn} revisit using body pose and object interaction features via a multi-stream model to fuse these high-level semantics with CNN features.
Martin~\textit{et al.}~\cite{martin2020dynamic_interaction} investigate modeling dynamic object interactions via graph neural networks for pose-based driver activity monitoring.
Tran~\textit{et al.}~\cite{tran2020realtime_detection_distracted} deploy a dual-camera system to detect multiple distracted driving behaviors by capturing body movements and face cues.
Zhao~\textit{et al.}~\cite{zhao2021driver} use adaptive spatial attention mechanism for driver activity detection.
More recently, Tan~\textit{et al.}~\cite{tan2021bidirectional} design a bidirectional posture-appearance interaction network to exploit RGB- and skeleton data in driver behavior recognition.
According to \cite{yang2021free}, latent space calibration helps to improve the classification performance of the deep learning approach for unseen class, only given few samples, which is hopefully to be leveraged to improve the performance of rarely-existed activity categories.
Differing from these CNN-based models, we put forward a vision-transformer-based framework to enhance both coarse- and fine-grained driver activity recognition with augmented feature distribution calibration inspired the approach leveraged for few shot classification~\cite{yang2021free}.

\noindent\textbf{Vision transformers.}
In modern times, transformer backbones have shown strong capacity in establishing long-range dependency information in image or video data~\cite{vaswani2017attention}, which prove beneficial for many downstream tasks.
Based on the pioneering work of Vision Transformer (ViT)~\cite{dosovitskiy2020vit} for image recognition, architectures of dense prediction transformers~\cite{liu2021swin,zhang2021trans4trans_t_its} and video classification transformers~\cite{liu2021video_swin,arnab2021vivit,wang2021efficient,peng2022proformer,chu2021twins} are created.
In the activity recognition area, Trear~\cite{li2021trear} proposes a transformer-based RGB-D egocentric activity recognition framework by adapting self-attention to model temporal structure from different modalities.
Besides, action-transformer~\cite{mazzia2021action_transformer}, motion-transformer~\cite{cheng2021motion}, hierarchical-transformer~\cite{cheng2021hierarchical}, spatial temporal transformer network~\cite{plizzari2021spatial} and STST~\cite{zhang2021stst} are designed for skeleton-based activity recognition, modeling temporal- and spatial dependencies in the skeleton sequences.
MM-ViT~\cite{chen2022mmvit} factorizes self-attention across the space, time, and modality dimensions, operating in the compressed video domain and exploiting various modalities.
Unlike these transformer methods, we design a feature calibration approach based on the feature extracted via vision transformers by interpolating features among different characterized clusters.

\section{Methods}
In this section, we introduce \textsc{TransDARC} -- a  \textsc{\textbf{Trans}}former-based framework for \textsc{\textbf{D}}river \textsc{\textbf{A}}ctivity \textsc{\textbf{R}}ecognition with Latent Space Feature \textsc{\textbf{C}}alibration shown by Fig.~\ref{fig:main}.
\textsc{TransDARC} is the first driver observation framework based on visual transformers and comprises a novel augmented feature distribution calibration module, which operates in the latent feature-space enriching and diversifying the training set at feature-level.
We first give a brief introduction of the basic components of vision transformer in Section~\ref{sec:trans}. Then, Section~\ref{sec:swin} provides a detailed description of  the leveraged Video Swin Transformer backbone~\cite{liu2021video_swin}. Finally, the most important contribution of our work, \textit{i.e.}, the complete \textsc{TransDARC} framework with augmented latent space feature calibration is introduced in Section~\ref{sec:transdarc}.
\subsection{Video transformer backbone}

\subsubsection{Vanilla vision transformer}
\label{sec:trans}
As introduced by~\cite{vaswani2017attention,dosovitskiy2020vit}, the vanilla transformer model is built based on stacks of the following basic components: Multi-head Self-Attention and Multi-Layer Perceptron together with Residual Connections and Layer Normalization respectively.
Before the image is passed to the transformer, it is divided into several patches to form the desired sequential input using a predefined fixed patch size.
A single Self-attention layer (SA), one of the most essential components inside the transformer block, comprises three basic elements: query $q$, key $k$ and value $v$, which can be calculated through the following equation:
$SA(q, k, v)=Softmax(qk^{T}/\sqrt{{\rho}_k})v$.
Note, that ${\rho}_k$ is a scaling factor aimed at avoiding the negative influence caused by the dot product of $q$ and $k$.
\begin{figure}[t]
\begin{center}
\includegraphics[width = 0.47\textwidth]{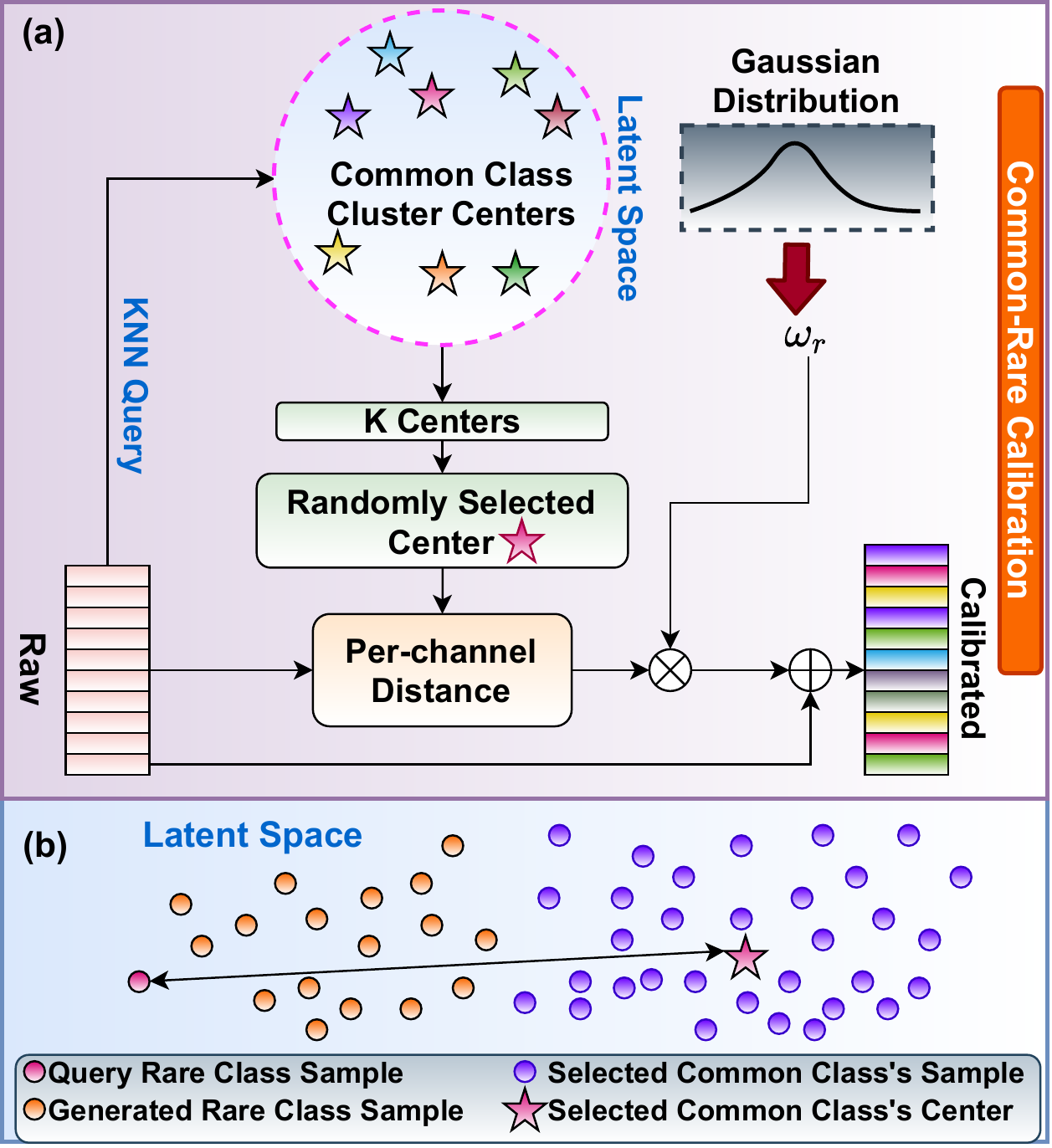}
\vskip-1ex
\caption{
Visualization of the common-rich feature calibration approach, where (a) indicates the workflow of the proposed common-rich calibration procedure and (b) indicates the visualization of generated feature in the latent space.
}
\vskip-5ex
\label{fig:commonrare}
\end{center}
\end{figure}
To obtain q, k, and v, linear projections $f$ are leveraged in SA and these three components are calculated following $q=f_q(I),~k=f_k(I)$, and $v=f_v(I)$, where $I$ indicates the input sequence of patches.
MSA connects multiple SA results and is computed through a concatenation of SA blocks represented as
$MSA(I)=Concat(SA_1, SA_2, ..., SA_N)f_{MSA}$. Several variants exist based on vision transformer, \textit{e.g.}, the Swin transformer~\cite{liu2021swin} utilizes a shifted window approach for patch embedding to obtain higher efficiency, which is utilized within the MSA layers (indicated by SW-MSA).

\subsubsection{Video Swin transformer}
\label{sec:swin}

We adopt the Video Swin transformer~\cite{liu2021video_swin} as our feature extraction backbone for driver activity recognition due to its excellent performance in traditional video classification.
As we deal with spatiotemporal video data, the MSA block is accompanied with 3D shifted window approach operating in time and space, as explained in~\cite{liu2021video_swin}.
Assuming  the input shape $[T, H, W]$ and the selected window size  $[K, N, N]$, $\frac{T}{K}\times\frac{H}{N}\times\frac{W}{N}$ patches are then extracted through window partition and the MSA is equipped with non-overlapping shifted window method
referred to as 3DW-MSA. 
Compared to the window position leveraged in the 3DW-MSA block, the configuration for window partition is shifted by $[\frac{K}{2}, \frac{N}{2}, \frac{N}{2}]$ along three axes with overlapping, denoted as 3DSW-MSA and later leveraged as the second MSA layer in the video Swin Transformer block.
The calculation procedure of the two consecutive Video Swin Transformer blocks can be represented according to~\cite{liu2021video_swin}:
\begin{equation}
\begin{aligned}
    \hat{z}^m &= DW-MSA(LN(z^{m-1})) + z^{m-1},\\
    z^m &=  FFN(LN(\hat{z^m}))+\hat{z}^m,\\
    \hat{z^{m+1}} &= 3DSW-MSA(LN(z^m))+z^m,\\
    z^{m+1} &= FFN(LN(\hat{z^{m+1}})) + \hat{z^{m+1}}.
\end{aligned}
\end{equation}
where $m$ indicates the $m$-th Video Swin Transformer block, $\hat{z}^m$ indicates the output of the 3DSW-MSA and FFN for block $m$, and FFN denotes the residual connection according to~\cite{liu2021video_swin}.  
In our framework, we first train Video Swin Transformer for our target task of driver behaviour understanding to obtain the feature extractor.
Then, intermediate Video Swin representations are used for the newly proposed data augmentations at the feature-level, which will now be described in detail.

\subsection{Latent space feature calibration}
\label{sec:transdarc}

After the pretraining of the Video Swin Transformer~\cite{liu2021video_swin} for driver activity recognition, we propose a novel latent space feature calibration method for further improvement of the recognition accuracy together with generalizability based on the embeddings from latent space extracted through discarding the last Fully-Connected (FC) layer of the Video Swin Transformer backbone.
Note, that we refer underrepresented categories as rare categories and overrepresented categories as common categories in our work.
Two groups of features are extracted in the first step: (1) features with the original video-based augmentations, \textit{i.e.}, using random video augmentation on the input video during inference, and (2) without original video-based augmentations. 
Let $V$ denote the raw video input and $T(\cdot)$ be the random augmentation procedure. 
Then, the resulting embedding $x$ and $x^*$ are calculated according to $x,~ x^* = M_{\theta}(V),~M_{\theta}(T(V))$.

The main goal of the latent space feature distribution calibration is to calibrate the rare classes, \textit{i.e.}, the activity categories containing  less samples in the training set according to a threshold $\eta$, based on the common existing classes in the latent space, in order to generate more features for such rare categories.
Note, that latent space feature calibration procedure is only executed on the training dataset to avoid looking at the validation and test sets and ensure fair comparisons.
Our approach is inspired by the feature calibration for few-shot recognition~\cite{yang2021free}, originally leveraged to extract more features based on the reference frames for the selected unseen classes.

\mypar{Statistics for the embedding in the latent space.} 
Following the assumption from \cite{yang2021free}, the feature distribution of each channel of the embedding in the latent space can be regarded as Gaussian distribution with the mean and co-variance, calculated according to the following equations:

\begin{equation}
\mu_i^j ,~\Sigma_i^j =\frac{\sum_{c=1}^{N_i}x_i^{j,c}}{N_c},~ \frac{1}{N_i-1}\sum_{c=1}^{N_i}(x_i^{j,c}-\mu_i^j)(x_i^{j,c}-\mu_i^j)^T,
\end{equation}
where $x_i^{j,c}$ indicates the jth channel of cth sample inside ith categories in the training embedding set and $N_i$ indicates the sample number of category $i$. 
\setlength{\textfloatsep}{5pt}
\begin{algorithm}[t]
    \caption{TransDARC -- latent space feature calibration}
    \label{algorithm}
    \renewcommand{\thealgorithm}{}
    \begin{algorithmic}[1]
        \small{
            \STATEx \textbf{Input:} $D_{train}$, $D_{test}$, and $D_{val}$ -- raw video training, testing, and evaluation sets; $N_{class}$ -- classes number; $K$ -- dominant category number; $\mu_{i}$ -- mean value for category i; $\Sigma$ -- covariance matrix for category i, $T(\cdot)$ -- random video augmentation; $H_{\gamma}$ -- attention-based classification head; $M_{\theta}$ -- well trained Video Swin Transformer with the last fully-connected layer discarded.
        }
        \STATE \textit{Initialize $L_{data} = [D_{train}, D_{test}, D_{test}]$}
        \STATE \% \textit{obtain random augmented raw video training dataset $D_{train}^*$}
        \STATE $D_{train}^* = T(D_{train})$
        \STATE \% \textit{Extract feature using $M_{\theta}$}
        \STATE $E_{train}$, $E_{test}$, $E_{val}$ = $M_{\theta}(D_{train})$, $M_{\theta}(D_{test})$, $M_{\theta}(D_{val})$
        \STATE $E_{train}^* = M_{\theta}(D_{train}^*)$
        \STATE \textit{Get class statistics $\left\{\mu_i \right\}_{i \in C}$, $\left\{\Sigma_i \right\}_{i \in C}$, $\left\{\mu_i^* \right\}_{i \in C}$, $\left\{\Sigma_i^* \right\}_{i \in C}$ regarding $E_{train}$ and $E_{train}^*$}
        \FORALL{\textit{action category $i$ and $i^*$ in $E_{train}$ and $E_{train}^*$}}
        \IF{\textit{$i$ and $i^*$ is a common activity category}}
        \State \textit{$E_{train}^{sc}$ and $E_{train}^{sc, *}$ $\leftarrow$ through self-augmented calibration}
        \ENDIF
         \IF{\textit{$i$ and $i^*$ are rare activity categories}}
        \State \textit{$E_{train}^{sc}$ and $E_{train}^{sc, *}$ $\leftarrow$ through rare-common feature calibration}
        \ENDIF
        \ENDFOR
        \State \textit{Form training feature set} $E_{train}^{calib} = [E_{train}^{sc}, E_{train}^{sc, *}, E_{train}^{sc}, E_{train}^{sc, *}]$
        \FOR{\textit{epoch $< N_{\max}$}}
        \STATE \textit{Training $H_{\gamma}$ based on $E_{train}^{calib}$}
        \ENDFOR
        \State \textit{Evaluate and test the performance of $H_{\gamma}$ on $E_{test}$, $E_{val}$.}
    \end{algorithmic}
\end{algorithm}
\begin{figure*}[t]

\begin{center}
\includegraphics[width = 1\textwidth]{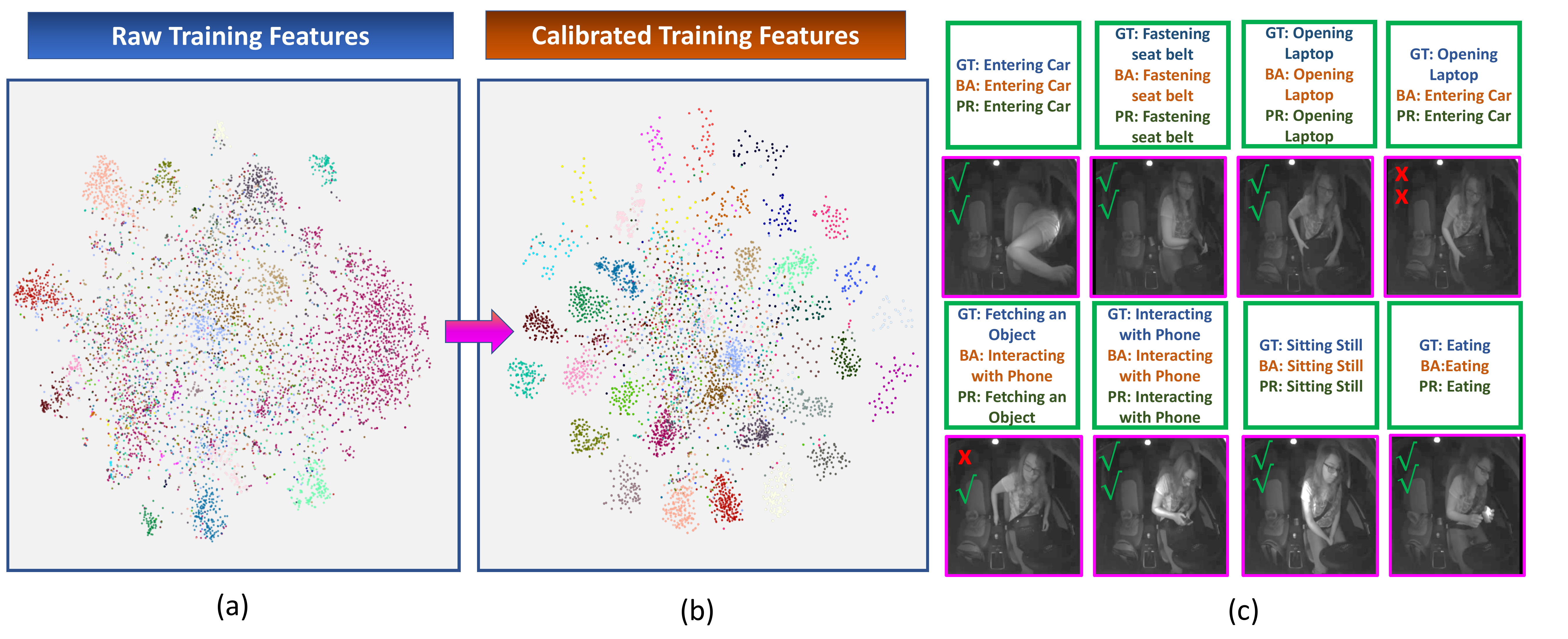}
\vskip -3ex
\caption{An illustration of the TSNE visualizations and the qualitative experimental results.
TSNE visualization of (a) the raw embeddings extracted from transformer-based feature extractor and (b) the generated embeddings through latent space feature calibration for fine-grained driver activity recognition. The training dataset of split0 in the driver activity recognition in fine-grained task is leveraged here.
In (c) GT indicates the groundtruth, BA indicates the prediction from Video Swin-Base baseline~\cite{liu2021video_swin} and PR indicates the prediction of our \textsc{TransDARC}.
}
\vskip -5ex
\label{fig:tsne}
\end{center}
\end{figure*} 

\mypar{Distribution calibration in rare-common wise.}
We first select the common activity categories denoted by $C_{com}$ as base classes according to predefined threshold $\eta$ and then calculate the aforementioned statistics for each base class in the latent space. Then $k$ closest categories, \textit{i.e.}, $C_{k}$, for each sample $x$ that belongs to the rare classes, are selected from $C_{com}$ according to the euclidean distance between $x$ and $\mu$. 
\begin{equation}
\mathbb{S}_{k} = Set(for ~j \in topk( \left \{ -\|\mu_i-x\| ~|~ i\in C_{com}\right\})),
\end{equation}
where $\mathbb{S}_{k}$ denotes the set of selected $k$ common categories. Then, instead of calculating the new Gaussian-like distribution for each channel of the embeddings in the latent space introduced in~\cite{yang2021free}, we leverage feature interpolation for distribution-based rare-category embedding generation.
For each feature sample $x_r$ of the rare class $r$, we first randomly select a category in $\mathbb{S}_k$ and then generate a random vector $\mathbf{\omega}_r$ which has the same channel size with $x_r$ and its dimension-wise values are sampled randomly through Gaussian distribution while ensuring the maximum absolute value as $1$.
Finally, the generated embedding in the latent space is indicated by the original feature softly augmented with the distance between the selected class center and itself, multiplied by the vector $\mathbf{\omega}_r$ to ensure random intensities of augmentation for different channels, which aims at using the top-k closest class centers to augment the selected embeddings in rare classes. 
We randomly choose $N_{rare}$ samples for each rare class based on the existing embeddings in the latent space during the execution of the rare-common feature generation procedure.
If the corresponding cluster center leveraged for calibration is ${\mu}_{c}$, then the generated new sample in the latent space can be represented by:
\begin{equation}
x_r^{new} = x_r + \omega_r \cdot D\left\{\mu_{c},~ x_r\right\},
\end{equation}
where $i\in [0, N_{rare}]$, and $D\left\{\cdot\right\}$ indicates the euclidean channel-wise distance in the latent space. For the features in the latent space generated with $T(\cdot)$, a similar procedure is leveraged as depicted in the following,
\begin{equation}
x_r^{*,new} = x_r^* + \omega_r^* \cdot D\left\{\mu_{c}^*,~ x_r^*\right\},
\end{equation}
where $\omega_r^*$ is the random vector generated for $x_r^*$.
The corresponding procedure can be also found in Fig.~\ref{fig:commonrare} for a detailed clarification.

\mypar{Distribution calibration for self-augment wise.}
Besides the feature calibration we have introduced for the rare classes, we simultaneously use the same augmented feature distribution calibration approach to generate more features for the common activity categories based on the its statistics.
To achieve this, we randomly generate new samples for common categories using top-k common categories to realize calibration based on the statistic characteristics defined by $\mu_i$ and $\Sigma_i$. 
Similar to the rare-common wise feature calibration, we first obtain the set of categories with the highest similarity score to $x_{c}$ as following, through random common category sampling.
\begin{equation}
\mathbb{Q}_{k} = Set(for ~j \in topk( \left \{ -\|\mu_i-x_{c}\| ~|~ i\in C_{c}\right\})).
\end{equation}
Then by given a random scaling factor $\omega_c$ and a random selected common category $\mu_s \in \mathbb{Q}_{k}$, the augmented sample can be obtained through,
\begin{equation}
x_{c}^{new} = x_{c} + \omega_c \cdot D\left\{\mu_{s},~ x_c\right\},
\end{equation}

Finally, $N_{com}$ samples for each common classes by repeating the aforementioned procedure several times.
We generate the features not only on the extracted features from the original rare input $I$, but also on the augmented input $T(I)$ as,
\begin{equation}
x_c^{*,new} = x_c^* + \omega_c^* \cdot D\left\{\mu_{s}^*,~ x_c^*\right\},
\end{equation}
while $x_c^{*,new}$ denotes the calibrated feature of the feature from augmented sample $x_c^{*}$ with radomly generated vector $\omega_r^{*}$.

The final latent space training set is composed of $E_{train}$ and $E_{train}^*$, and each of them is composed of the generated common category samples and rare category samples.
We then train the attention-based recognition head instead of a single FC layer, as depicted on the right hand side of Fig.~\ref{fig:main}, denoted as $H_{\gamma}$ for $N_{max}$ epochs in total.

\mypar{Hard sample mining.} With a predefined epoch frequency $N_{mine}$, we run the inference model to obtain sample-wise loss and select the hard samples with the loss higher than $\delta$ times of the mean loss in the training set.
Then, we simply train $N_{hard}$ epochs based on these hard samples for the attention-based driver activity recognition head $H_{\gamma}$ depicted by Fig.~\ref{fig:main}, aiming to strengthen the supervision on the difficult samples. The workflow of the proposed \textsc{TransDARC} pipeline is illustrated in detail in Alg.~\ref{algorithm}.

\section{Experiments}

\begin{table*}[t]
\caption{An overview of the experimental results on Drive\&Act~\cite{martin2019drive} for multiple tasks, including coase- and fine-grained driver activity recognition and human-object interaction triplet recognition. The evaluation metric is Top-1 Accuracy.}
\vskip-2ex

\centering
\scalebox{0.8}{\begin{tabular}{llllll|l|lllllllllll}
\toprule
\multicolumn{1}{c}{\multirow{2}{*}{\begin{tabular}[c]{@{}c@{}}\\\textbf{Model }\end{tabular}}} & \multicolumn{2}{c}{{\cellcolor[rgb]{0,0.502,0.502}}\textbf{\textcolor{white}{Fine-grained}}} & \multicolumn{1}{l|}{} & \multicolumn{2}{c|}{{\cellcolor[rgb]{0,0.502,0}}\textbf{\textcolor{white}{Coarse task}}} &  & \multicolumn{2}{c}{{\cellcolor[rgb]{0,0.314,0.682}}\textbf{\textcolor{white}{Action}}} &  & \multicolumn{2}{c}{{\cellcolor[rgb]{1,0.427,0.412}}\textbf{\textcolor{white}{Object}}} &  & \multicolumn{2}{c}{{\cellcolor[rgb]{0.89,0.435,0.58}}\textbf{\textcolor{white}{Location }}} &  & \multicolumn{2}{c}{{\cellcolor[rgb]{0.412,0.647,0.753}}\textbf{\textcolor{white}{All}}} \\
\multicolumn{1}{c}{} & \multicolumn{1}{c}{Val} & \multicolumn{1}{c}{Test} &  & \multicolumn{1}{c}{Val} & \multicolumn{1}{c|}{Test} &  & \multicolumn{1}{c}{Val} & \multicolumn{1}{c}{Test} &  & \multicolumn{1}{c}{Val} & \multicolumn{1}{c}{Test} &  & \multicolumn{1}{c}{Val} & \multicolumn{1}{c}{Test} &  & \multicolumn{1}{c}{Val} & \multicolumn{1}{c}{Test} \\ 
\midrule
\multicolumn{18}{c}{\textbf{Previously published approaches}} \\ 
\midrule
Pose~\cite{martin2019drive} & 55.17 & 44.36 &  & 37.18 & 32.96 &  & 57.62 & 47.74 &  & 51.45 & 41.72 &  & 53.31 & 52.64 &  & 9.18 & 7.07 \\
Interior~\cite{martin2019drive} & 45.23 & 40.30 &  & 35.76 & 29.75 &  & 54.23 & 49.03 &  & 49.90 & 40.73 &  & 53.76 & 53.33 &  & 8.76 & 6.85 \\
2-Stream~\cite{martin2019drive} & 53.79 & 45.39 &  & 39.37 & 34.81 &  & 57.86 & 48.83 &  & 52.72 & 42.79 &  & 53.99 & 54.73 &  & 10.31 & 7.11 \\
3-Stream~\cite{martin2019drive} & 55.67 & 46.95 &  & 41.70 & 35.45 &  & 59.29 & 50.65 &  & 55.59 & 45.25 &  & 59.94 & 56.50 &  & 11.57 & 8.09 \\
C3D~\cite{tran2015learning} & 49.54 & 43.41 &  & - & - &  & - & - &  & - & - &  & - & - &  & - & - \\
P3D~\cite{qiu2017learning} & 55.04 & 45.32 &  & - & - &  & - & - &  & - & - &  & - & - &  & - & - \\
I3D~\cite{carreira2017quo} & 69.57 & 63.64 &  & 44.66 & 31.80 &  & 62.81 & 56.07 &  & 61.81 & 56.15 &  & 47.70 & 51.12 &  & 15.56 & 12.12 \\
CTA-NET~\cite{wharton2021coarse} & 72.42 & 65.25 &  & 62.82 & 52.31 &  & 57.59 & 56.41 &  & 63.37 & 59.19 &  & 56.41 & 63.01 &  & 46.44 & 49.41 \\ 
\midrule
\multicolumn{18}{c}{\textbf{Our framework}} \\ 
\midrule
\rowcolor[rgb]{0.753,0.753,0.753}  Video Swin~\cite{liu2021video_swin} & 88.10 & 85.74 &  & 82.67 & 78.53 &  & 92.60 & 91.32 &  & 89.10 & 86.38 &  & 85.74 & 85.48 &  & 85.74 & 85.48 \\
\rowcolor[rgb]{0.753,0.753,0.753} \textsc{TransDARC} (ours) & \textbf{93.58} & \textbf{89.65} &  & \textbf{83.42} & \textbf{79.69} &  & \textbf{93.86} & \textbf{92.54} &  & \textbf{90.70} & \textbf{87.19} &  & \textbf{87.59} & \textbf{86.99} &  & \textbf{87.44} & \textbf{86.97} \\
\bottomrule
\end{tabular}}
\vskip-3ex
\label{tab:main}
\end{table*}

\begin{table}[t]
\centering
\caption{An overview of the performance evaluation for rarely and commonly existed categories for fine-grained driver activity recognition.}
\vskip-2ex
\label{tab:rare_rich_eval}
\scalebox{0.8}{\begin{tabular}{lllll} 
\toprule
\multicolumn{1}{c}{\textbf{Model }} && \multicolumn{1}{c}{\textbf{Common}} & \multicolumn{1}{c}{\textbf{Rare}} & \multicolumn{1}{c}{\textbf{All}} \\ 
\hline
\multicolumn{5}{c}{\textbf{Validation for existing framework}} \\ 
\midrule
C3D~\cite{tran2015learning,roitberg2020cnn_spatialtemporal} (implemented by~\cite{roitberg2020cnn_spatialtemporal}) && 54.44 & 45.70 & 50.07 \\
Pseudo 3D ResNet~\cite{qiu2017learning,roitberg2020cnn_spatialtemporal} && 58.00 & 52.08 & 55.04 \\
I3D~\cite{carreira2017quo,roitberg2020cnn_spatialtemporal} && 80.62 & 58.50 & 69.67 \\ 
\midrule
\multicolumn{5}{c}{\textbf{Validation for our framework}} \\ 
\midrule
Video Swin baseline~\cite{liu2021video_swin} && 88.57 & 82.39 & 85.74 \\
Ours && \textbf{94.41} & \textbf{83.44} & \textbf{93.42} \\ 
\midrule
\multicolumn{5}{c}{\textbf{Test for existing approaches}} \\ 
\midrule
C3D~\cite{tran2015learning,roitberg2020cnn_spatialtemporal}  && 47.97 & 38.86 & 43.41 \\
Pseudo 3D ResNet~\cite{qiu2017learning,roitberg2020cnn_spatialtemporal} && 52.43 & 38.20 & 45.32 \\
I3D~\cite{carreira2017quo,roitberg2020cnn_spatialtemporal} && 77.88 & 49.41 & 63.64 \\ 
\midrule
\multicolumn{5}{c}{\textbf{Test for our approaches}} \\ 
\midrule
Video Swin baseline~\cite{liu2021video_swin} && 86.65 & 76.45 & 85.74 \\
Ours && \textbf{90.83} & \textbf{77.55} & \textbf{89.65} \\
\bottomrule
\end{tabular}}
\end{table}

\begin{table}[t]
\centering
\caption{An overview of the experimental results for cross-modality evaluation while trained on NIR\_1.}
\vskip-2ex
\scalebox{0.66}{\begin{tabular}{lllllllll} 
\toprule
\textbf{Model} & \textbf{NIR\_1} & \textbf{NIR\_2} & \textbf{NIR\_3} & \textbf{NIR\_4} & \textbf{NIR\_5} & \textbf{K\_color} & \textbf{K\_depth} & \textbf{K\_ir} \\ 
\midrule
I3D~\cite{carreira2017quo} & 69.57 & 4.51 & 6.96 & 7.39 & 9.03 & 5.41 & 3.00 & 5.77 \\
Video Swin~\cite{liu2021video_swin} & 88.10 & 5.60 & 7.50 & 9.90 & 13.70 & 8.63 & 6.13 & 8.62 \\
Ours & \textbf{93.58} & \textbf{9.11} & \textbf{11.50} & \textbf{20.87} & \textbf{16.40} & \textbf{12.10} & \textbf{8.01} & \textbf{11.21} \\
\bottomrule
\end{tabular}}

\label{tab:modality}
\end{table}
\begin{table}[t]
\centering
\caption{Experimental results for different classification heads.}
\vskip-2ex
\scalebox{0.7}{\begin{tabular}{lllll} 
\toprule
 & \multicolumn{2}{c}{\textbf{Fine-grained}} & \multicolumn{2}{c}{\textbf{Coarse}} \\
\textbf{Head} & \multicolumn{1}{c}{\textbf{val}} & \multicolumn{1}{c}{\textbf{test}} & \multicolumn{1}{c}{\textbf{val}} & \multicolumn{1}{c}{\textbf{test}} \\ 
\hline
fully connected & 91.98 & 88.59 & 82.68 & 79.55 \\
Ours & \textbf{93.58} & \textbf{89.65} & \textbf{83.42} & \textbf{79.69} \\
\bottomrule
\end{tabular}}
\label{tab:head}
\end{table}

\begin{table}
\centering
\caption{Ablation regarding different augmentation components.}
\vskip-3ex

\label{tbl:multi-abl}
\scalebox{0.7}{\begin{tabular}{lcc} 
\toprule
\multirow{2}{*}{\textbf{Experiments}} & \multicolumn{2}{l}{\textbf{Fine-grained Split0}} \\
 & \textbf{Val} & \textbf{Test} \\
 \midrule
Video Swin~\cite{liu2021video_swin} & 90.60 & 86.36 \\
Calibration with only Augmentation from~\cite{yang2021free} & 91.92 & 86.40\\
\midrule
\textsc{TransDARC} without Self-Augment & 92.56 & 87.83 \\
\textsc{TransDARC} without augmented feature calibration & 92.71 & 87.70 \\
\textsc{TransDARC} & \textbf{93.47} & \textbf{87.83} \\
\bottomrule
\end{tabular}}
\end{table}

 \subsection{Dataset}
\mypar{Drive\&Act dataset} Drive\&Act~\cite{martin2019drive} is the largest public driver observation  dataset targeting  both coarse- and fine-grained driver activity recognition  and covering  $12$ hours (over $9.6$ million frames) of  distracted driving recordings inside the vehicle. RGB, infrared, depth and 3D skeleton data collected from six different views are provided in the dataset.
The  videos are hierarchically annotated resulting in $83$ different driver behaviour categories in total.
Drive\&Act contains $3$ splits for training and evaluation (with no driver overlap between the training, validation and test sets), which we adopt to keep fair comparisons to previous works.
The results of the three validation and test sets are averaged.
The leveraged different sensors in our work are marked as NIR\_1, NIR\_2, NIR\_3, NIR\_4, NIR\_5, K\_color, K\_depth, and K\_ir, indicating the NIR Front-top, NIR Right-top, NIR Back, NIR Face-view, NIR Left-Top, Kinect RGB, Kinect Depth, and Kinect IR modalities~\cite{martin2019drive}, respectively. 
\subsection{Implementation Details}
We use Video Swin Base as our feature extraction backbone which is trained on a Quatumn 8000 graphic card with a batch-size $4$ for $22$ epochs using initial learning rate as $1e^{-4}$, AdamW~\cite{loshchilov2017decoupled} optimizer and cosine annealing learning rate scheduler. 
The Video Swin Base backbone is implemented into driver activity recognition in our task by selected two clips containing $32$ frames individually based on the raw video input with step size $2$ and randomly initialized start-frame-index. 
For the fine-grained task, $N_{mine}$, $\delta$, $\eta$ and $N_{hard}$ are set to $30$, $1.2$, $400$ and $1$ and the attention-based classification head is optimized for $1200$ epochs. 
The feature of these two video clips selected with fixed temporal step and random start time points are averaged and then fed into the FC layer for classification.
Consistent with previous work~\cite{martin2019drive,wharton2021coarse,roitberg2020cnn_spatialtemporal}, we use balanced accuracy (average per-class accuracy) as our main evaluation metric. 
More details regarding the hyper parameters is provided in our code.

\subsection{Analyses}
\mypar{Does TransDARC perform well on driver activity recognition?}
Our extensive experiments on Drive\&Act showcase that the  answer is definitely yes. 
In Table~\ref{tab:main}, the previously published driver observation methods are grouped in the existing approaches block and the performances of Video Swin Transformer~\cite{liu2021video_swin}, adapted to the task of driver observation by us as well as the proposed \textsc{TransDARC} approach are represented in the lower block.
The Drive\&Act~\cite{martin2019drive} dataset distinguishes between three recognition tasks: fine-grained driver activities (which is selected as the main evaluation mode~\cite{martin2019drive}), recognition of \textit{coarser} driver behaviours, (\textit{i.e.}, the long-term tasks the person wants to accomplish) and recognition of more primitive human-object interactions represented as action-object-location triplets.
First, we consider the fine-grained driver activity recognition evaluation which is used as the  main evaluation level in the past~\cite{martin2019drive}.
We observe a significant gain in accuracy using a transformer-based  backbone (Video Swin) alone, \textit{i.e.}, $15.68\%$ on the validation set and $20.49\%$ on the Drive\&Act test set (Table~\ref{tab:main}).
This impressive performance boost provides encouraging evidence, that driver behaviour models could benefit more from the novel developments of attention-based neural networks for feature extraction~\cite{vaswani2017attention}.
The best recognition results are consistently achieved withe the complete \textsc{TransDARC} framework with latent space enhancement.
\textsc{TransDARC} outperforms the best previously published driver activity recognition approach~\cite{wharton2021coarse} by $21.16\%$ and $24.40\%$ and the Video Swin-only baseline without latent space calibration by $5.48\%$ and $3.91\%$ on the validation and test set respectively.
Furthermore, experiments are conducted in Table~\ref{tbl:multi-abl} to indicate the efficacy of each augmentation component.
This validates that feature augmentations conducted in \textsc{TransDARC}  indeed improve the embedding quality presumably by mitigating issues limiting the driver observation quality, \textit{e.g.}, the unbalanced class distribution and the implicitly existing features that cannot be fully explored by conventional data augmentation on the raw videos.
Fig.~\ref{fig:tsne} provides a 2-dimensional embedding of the training features before and after the latent space calibration using t-SNE Fig~\ref{fig:tsne} step (both using the same backbone).
The category boundaries of the \textsc{TransDARC} features (represented by (b)) depict a much better boundary localization quality and concise clusters  compared to the raw training features (represented by (a)). 
Lastly, in Fig.~\ref{fig:tsne}(c) we showcase  quantitative prediction results for fine-grained driver behaviour recognition, illustrating the effectiveness of the proposed \textsc{TransDARC} method.
We also conduct experiments to evaluate the performance of the proposed attention-based head for driver activity recognition according to Table~\ref{tab:head} as described in Fig.~\ref{fig:main} and Alg.~\ref{algorithm}, illustrating clear benefits of the attention-based in contrast to a  fully-connected layer for fine-grained  driver activity recognition.
 
\mypar{How about the cross-task generalization of TransDARC?}
Next, we investigate the quality of \textsc{TransDARC} predictions at different levels of driver behaviour granularities (Table ~\ref{tab:main}).
Our  approach outperforms the Video Swin-only baseline and all the previously published existing approaches considered on this benchmark by a significant margin and setting a new state-of-the-art performance not across all granularity levels:  fine-grained driver activities, coarse driver behaviours and primitive human-object interactions (marked with triplets of action, object and location).
In the last column of Table ~\ref{tab:main}  the combined accuracy of these three items, \textit{i.e.}, action, object and location,  is reported as the proportion of the predictions, where \textit{all} these components were recognized correctly.
Consistently across all granularity levels, the Top-1 accuracies are improved by both, using the transformer-based backbone and the complete  \textsc{TransDARC} framework with feature calibration.

\mypar{Does TransDARC generalize well to novel sensors and modalities?}
According to our cross-modal recognition experiments, the answer is clearly yes. 
Cross-modal generalization and robustness to domain shifts  is  essential for deep learning-based ADAS systems, since sensors locations depends on the cabin structure  and differ from vehicle-to vehicle~\cite{reiss2020deep}. 
On the other hand, re-training a deep learning model again and again if  the position has changed is costly and time consuming. 
In Table~\ref{tab:modality}, we evaluate the performance of our \textsc{TransDARC} approach trained exclusively on the NIR\_1 view on all $7$  Drive\&Act modalities, of which 6 have never been seen during training.
Since our augmentations enrich the  training data at feature-level, it is not surprising that \textsc{TransDARC} leads to a large gain in accuracy compared with the Video Swin transformer baseline and the convolutional I3D approach without such feature calibration. 
This effect is especially large for NIR\_4, where \textsc{TransDARC} outperforms Video Swin and I3D by $10.97\%$ and $13.48\%$ for the fine-grained driver activity recognition on the validation set respectively, highlighting the quality of the proposed feature space calibration under cross-modal conditions.

\mypar{Performance of TransDARC on common and rare driver behaviours in an unbalanced dataset.} 
Next, we report the Top-1 accuracy for fine-grained driver activities which are over- and underrepresented in the training set separately.
We follow the evaluation protocol of~\cite{roitberg2020cnn_spatialtemporal} and use the terms \textit{common} and \textit{overrepresented} as well as \textit{rare} and \textit{underrepresented} interchangeably.
We compare \textsc{TransDARC} to the Video Swin Transformer-only baseline, and CNN-based approaches, \textit{i.e.}, C3D, P3D, and I3D evaluated in~\cite{roitberg2020cnn_spatialtemporal}.
From the perspective of \textit{rare} driver activity categories, \textit{e.g.}, \textit{closing laptop}, \textit{opening backpack} and \textit{putting on sunglasses} the Video Swin Transformer baseline~\cite{liu2021video_swin} adapted by us for driver observation surpasses all other models by a considerable margin, e.g., surpassing I3D  by $23.89\%$ and $27.04\%$ on the validation and test sets respectively.
The complete \textsc{TransDARC} framework shows a further performance improvement by $1.05\%$ and $1.10\%$  for \textit{rare} driver activities.
Surprisingly, there is largeer performance improvement brought by \textsc{TransDARC} on the \textit{common}  driver behaviours, \textit{e.g.}, \textit{sitting still}, \textit{eating} and \textit{interacting with phone}. 
The underlying reason for this is presumably due to the quality of the generated features in the latent space being  dependent on the training samples diversity.
Video Swin Transformer outperforms I3D on the for \textit{common} activities by $7.95\%$ and $8.77\%$, while  \textsc{TransDARC} further improves  performance  by $5.84\%$ and $4.18\%$ on the validation and test sets respectively.

\section{Conclusion}

In this work, we introduced \textsc{TransDARC} -- a novel approach for identifying driver secondary activities in video.
Our approach for the first time leverages a visual transformer backbone for  driver monitoring and enhances this model with a novel augmented feature distribution calibration module which diversifies the training set at feature-level therefore facilitating generalization to novel data appearances under cross-modal and cross-view conditions.
Our framework achieves state-of-the-art performance on all tasks of the challenging Drive\&Act benchmark, including fine-grained and coarse driver activity recognition as well as  human-object interaction detection inside the vehicle.
Our experiments clearly indicate that the proposed feature calibration module indeed improves the latent space feature set, which is validated quantitatively on a public benchmark and qualitatively via cluster analysis.
Overall, our framework provides a way for more accurate and well-generalizable ADAS systems and will also be considered for other  tasks, such as recognition of daily living activities in household robotics, in the future.

\bibliographystyle{IEEEtran}
\bibliography{egbib}

\end{document}